\documentclass[letterpaper, 10 pt, conference]{ieeeconf}  

\IEEEoverridecommandlockouts                              

\usepackage{amsmath} 
\DeclareMathOperator*{\argmin}{argmin}
\DeclareMathOperator*{\argmax}{argmax}

\DeclareMathOperator{\diag}{diag}
\usepackage{amssymb}  
\usepackage{graphicx}
\usepackage{color,xcolor,soul}
\usepackage{cite}
\usepackage{physics}
\usepackage{amsthm}
\usepackage{extra_commands}
\usepackage{algorithm, algorithmic}
\usepackage{multirow}
\usepackage{mathtools}
\usepackage{hyperref}
\usepackage{siunitx}
\usepackage{makecell}

\newtheorem{prop}{Proposition}
\newtheorem{remark}{Remark}
\theoremstyle{definition}
\newtheorem{definition}{Definition}

\newcommand{\figureFolder}{figures}

\title{\LARGE \bf
Approximate Optimal Controller Synthesis for Cart-Poles and Quadrotors via Sums-of-Squares
}

\author{Lujie Yang$^{1}$, Hongkai Dai$^{2}$, Alexandre Amice$^{1}$, Russ Tedrake$^{1,2}$
\thanks{$^{1}$ Massachusetts Institute of Technology}%
\thanks{$^{2}$ Toyota Research Institute }%
\thanks{{Emails: \tt\small\{lujie, amice, russt\}@mit.edu,  hongkai.dai@tri.global}}
}

\begin{document}
\maketitle
\thispagestyle{empty}
\pagestyle{empty}

\begin{abstract}
Sums-of-squares (SOS) optimization is a promising tool to synthesize certifiable controllers for nonlinear dynamical systems. Building upon prior works \cite{lasserre2008nonlinear, jiang2015global}, we demonstrate that SOS can synthesize dynamic controllers with bounded suboptimal performance for various underactuated robotic systems by finding good approximations of the value function. We summarize a unified SOS framework to synthesize both \emph{under-} and \emph{over-} approximations of the value function for continuous-time, control-affine systems,  use these approximations to generate approximate optimal controllers, and perform regional analysis on the closed-loop system driven by these controllers. We then extend the formulation to handle hybrid systems with contacts. We demonstrate that our method can generate tight under- and over- approximations of the value function with low-degree polynomials, which are used to provide stabilizing controllers for continuous-time systems including the inverted pendulum, the cart-pole, and the quadrotor as well as a hybrid system, the planar pusher. To the best of our knowledge, this is the first time that a SOS-based time-invariant controller can swing up and stabilize a cart-pole, and push the planar slider to the desired pose. Videos at \href{https://youtu.be/QQR_pPNPeyg}{https://youtu.be/QQR\_pPNPeyg}; demo code at \href{https://deepnote.com/workspace/lujieyang-d88573a8-1f76-410d-81bf-e0d65a047c67/project/hjb-sos-7a7b3bf1-fa02-4814-9b91-7df510a7f674/\%2Fcubic.ipynb}{https://deepnote.com/workspace/lujieyang/project/hjb-sos}.
\end{abstract}

\section{INTRODUCTION}
Many interesting robotic tasks, including running, flying, and manipulation are naturally formulated as optimal control problems. Dynamic programming \cite{bellman1966dynamic, bertsekas2012dynamic}  
and the Hamilton-Jacobi-Bellman (HJB) equation provide a sound theoretical framework to study the solution to such problems and obtain optimal controllers to the nonlinear dynamical systems. However, the HJB equation is challenging to solve and most methods suffer from an exponential growth of computational complexity, widely recognized as the ``curse of dimensionality" \cite{bellman2015applied}. This challenge has motivated the study of numerically tractable methods to provide approximate solutions to the HJB equation \cite{de2003linear, tedrake2009underactuated}.

Reinforcement learning has demonstrated great empirical success in finding approximate solutions to the HJB equation which in turn generate highly dynamic controllers. Despite their great empirical success, such methods are rarely, if ever, able to bound the suboptimality of the resulting controllers or the approximation quality of the true cost-to-go function due to their reliance on finite sampling and function approximators.

These drawbacks have motivated the study of a variety of methods that provide such guarantees. 
In \cite{summers2013approximate, wang2015approximate}, the authors lower bound the value function (the viscosity solution \cite{crandall1984some, crandall1983viscosity} to the HJB partial differential equation) by relaxing the HJB equation to an inequality and only consider discrete-time systems. 
Other works \cite{lasserre2008nonlinear} consider continuous-time systems and provide under-approximation of the value function based on the moment-SOS hierarchy; this forms the basis of our approach. We show that with careful attention to a few details, their results can be extended to the interesting dynamic regimes. 
On the other hand, \cite{jiang2015global} relaxes the optimal control problem to find a global over-approximator on the value function using a policy iteration method. We extend this work by explicitly removing the assumption of an initial, globally stabilizing,  polynomial control law,
which may not exist even for polynomial dynamics.
\begin{figure}
\centering
	\includegraphics[width=0.47\textwidth]{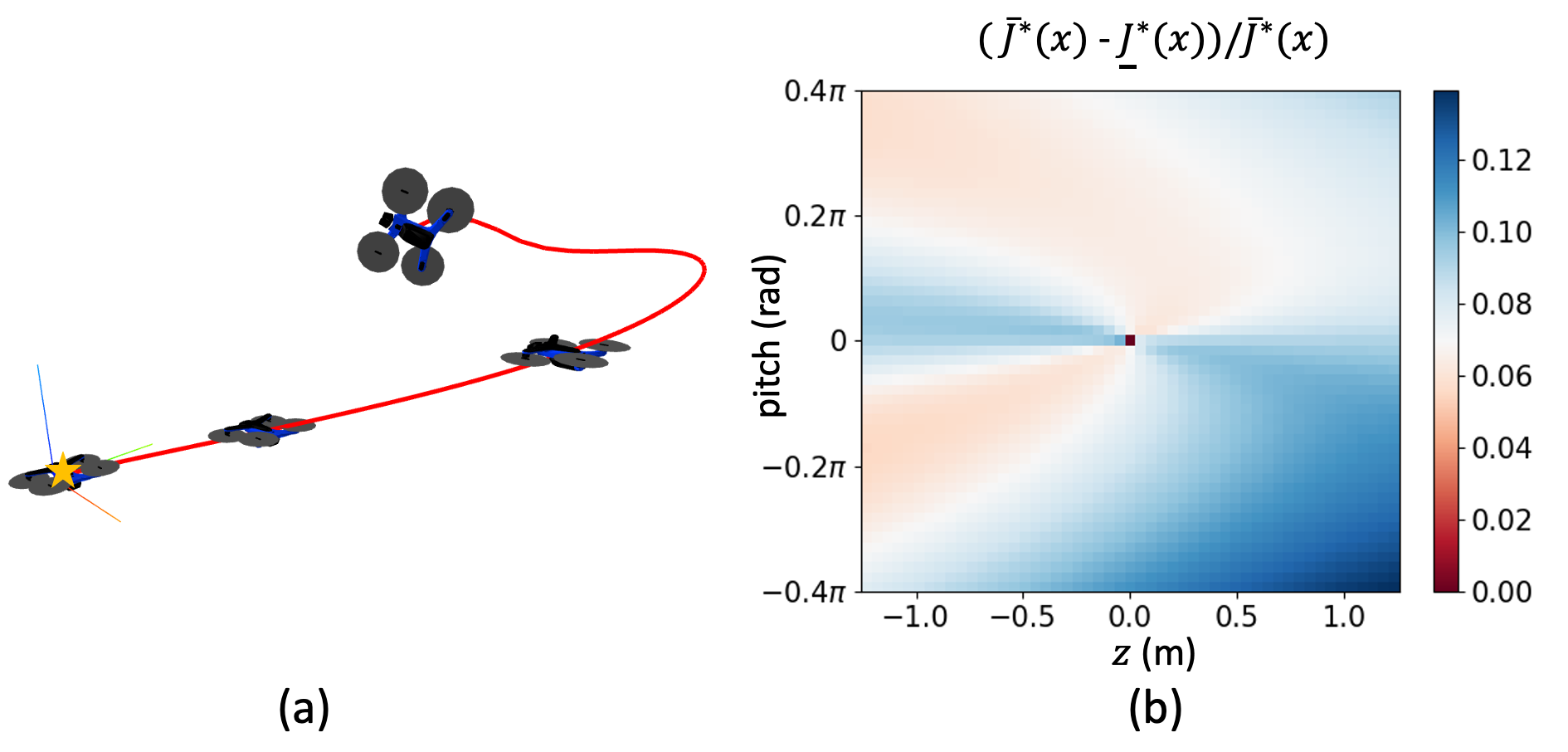}
	\caption{Quadrotor: (a) snapshots of the simulation using the stabilizing controller synthesized from value function approximations. The red curve is the quadrotor's trajectory to balance around the origin. \href{https://hongkai-dai.github.io/figures/quadrotor3d.html}{Here} is an animation. (b) $(z, $ pitch) slice of the relative approximation error of the value function under- and over- estimates ($\underline J^*$ and $\bar J^*$ respectively).} 
	\label{fig:quadrotor3d_banner}
	\vspace*{-0.68cm}
\end{figure}

In this work, we provide a unified framework to obtain both an under- and over- approximation of the value function within a compact region for continuous-time nonlinear systems. We synthesize tight value function approximations (see Fig. \ref{fig:quadrotor3d_banner}) satisfying relaxed HJB conditions using convex optimization, specifically sums-of-squares (SOS) programming \cite{parrilo2000structured}, which can be solved in polynomial time.
The approximation is generated locally over regions of interest for various underactuated robotic systems, which improves the approximation quality when compared to many works seeking global approximators. This is due to the fact that the SOS conditions are stronger \cite{schmudgen2017moment} over a compact set than over unbounded sets.

While a significant body of work in the controls literature \cite{prajna2004nonlinear,prajna2004nonlinearconvex} has adopted SOS-based methods for verifying and analyzing the stability and safety of nonlinear systems, their focus has primarily been on Lyapunov or barrier certificates. SOS has been used in the robotics community to cover the state-space with stabilizing \cite{tedrake2010lqr, majumdar2016funnel} and safe \cite{clark2021verification, wang2022safety} controllers. These works emphasize stability or safety rather than the optimality of the local controller and consider dynamics locally approximated by their Taylor expansion. In this work, we show that using the original robot dynamics without local approximations and incorporating a notion of optimality can dramatically increase the size of the regions over which the resulting SOS-based controllers stabilize the system. Moreover, these prior works require a locally stabilizing initial controller for non-autonomous systems. In contrast, our approach does not require any initial guess on the controller to synthesize a value function under-approximator, which can be used to derive a stabilizing controller for synthesizing a value function over-approximator in all of our experiments. 

Our contributions are summarized as follows: 1) we present a strengthened numerical relaxation of the programs from \cite{tedrake2009underactuated, jiang2015global} for computing value function estimates that approximately satisfy the HJB over a compact domain. 2) We analyze the local performance of our value approximations by computing inner approximations of both the closed-loop system's region of attraction and the region in which the controllers synthesized from our value function approximations are guaranteed to perform well. 
3) We apply our approach to continuous robotic systems such as the inverted pendulum, cart-pole, and quadrotor. We showcase that tight under- and over- estimates of the value function can be synthesized and the corresponding controller derived from these approximators can stabilize the systems for a large region of the state space. 
4) We extend the under-approximation formulation to hybrid systems with contacts and validate the framework on the hybrid planar-pusher system. To the best of our knowledge, our method provides the first time-invariant polynomial controllers synthesized with SOS to achieve full cart-pole swing-up and accomplish the planar-pushing task.

\section{PROBLEM STATEMENT}
We consider a continuous-time control-affine system:
\begin{align}
    \dot{x}(t) = f(x(t), u(t)) = f_1(x(t)) + f_2(x(t))u(t),
    \label{eq:dynamics}
\end{align}
where $f_1$ and $f_2$ are polynomial functions of $x(t) \in \mathbb{R}^{n_x}$ and $u(t) \in \mathcal{U} = \{u|u_{\text{min}} \le u \le u_{\text{max}}\} \subset \mathbb{R}^{n_u}$ is the control input.

We consider an optimal control problem of regulating the system to a desired set point $x_d \in \mathbb{R}^{n_x}$ incurring the instantaneous cost
\begin{align*}
    l(x, u) = q(x-x_d) + u^T R u,
\end{align*}
where $q(\bullet)$ is a positive-definite polynomial and $R$ is a diagonal matrix with positive entries. Under the system dynamics \eqref{eq:dynamics}, the optimal cost-to-go function is positive definite and defined as
\begin{equation}
\begin{aligned} \label{eq:cost_to_go} 
    J^*(x_0) &= \min_{u \in \mathcal{U}}\int^{\infty}_0 l(x(t), u(t)) dt,  \; x(0) = x_0,
\end{aligned}
\end{equation}
while the cost-to-go under a policy $\pi$ is defined as
\begin{align}
J^{\pi}(x_0) = \int_0^{\infty} l(x(t), \pi(x(t))) \; dt, \; x(0) = x_0.
\end{align}
The Hamilton-Jacobi-Bellman equation defines a necessary and sufficient condition for a function $J^*$ to be the optimal cost-to-go \cite{crandall1983viscosity}:
\begin{align}
    \forall x,\; \min_{u \in \mathcal{U}} \biggl[l(x, u) + \frac{\partial J^*}{\partial x}f(x, u)\biggr] = 0.
    \label{eq:hjb_eqn}
\end{align}
Using \eqref{eq:hjb_eqn}, the optimal controller can be given in closed-form for the case of diagonal $R$ (proof in Appendix \ref{subsection:optimal_controller_proof}):
\begin{align}
    \pi^*(x) &=  \Tilde \pi(x; J^*)
    \label{eq: minimizing_controller_closed_form} \\
    \Tilde \pi(x; J) &\coloneqq \argmin_{u \in \mathcal{U}} \biggl[l(x, u) + \frac{\partial J}{\partial x}f(x, u)\biggr] \label{eq:controller_qp} \\
    &= \text{clamp}\left(-\frac{1}{2}R^{-1}f_2(x)^T \frac{\partial{J}}{\partial x}^T, u_{\text{min}}, u_{\text{max}}\right), \label{eq: minimizing_controller_closed_form_general}
\end{align}
where $\text{clamp}$ function is defined as $\text{clamp}(u, u_{\text{min}}, u_{\text{max}}) = \min(\max(u, u_{\text{min}}), u_{\text{max}})$ elementwise. 

In general, it is intractable to find an analytic solution $J^*(x)$ to the HJB equation\cite{powell2007approximate}, so we aim to find good approximations of the value function, by providing both under- and over-approximators for $J^*(x)$.

\subsection{Global value function under-approximation}
It is shown in \cite{rantzer2000piecewise} that $\underline J$ is a \textit{global} under-estimate of the value function if it satisfies:
\begin{align}
    \forall x,\; \min_{u \in \mathcal{U}} \biggl[l(x, u) + \frac{\partial \underline{J}}{\partial x}f(x, u)\biggr] \ge 0. \label{eq:HJB_ineq_lo}
\end{align}
This under-estimate HJB inequality is equivalent to 
\begin{align}
    \forall x, \forall u \in \mathcal{U}, \; l(x, u) + \frac{\partial \underline{J}}{\partial x} f(x, u) \ge 0.
    \label{eq: J_lower_inequality}
\end{align}
A good under-approximator can be found using the optimization program:
\begin{subequations}
\label{eq:J_lower_program}
\begin{equation} \label{eq:J_lower_program_a}
    \max \int_{\mathcal{X}} \underline{J}(x)\; dx
\end{equation}
\begin{equation}\label{eq:J_lower_program_b}
    \text{s.t. } \; 
    l(x, u) + \frac{\partial \underline{J}}{\partial x} f(x, u) \geq 0, \quad \forall x, \forall u \in \mathcal{U}, 
\end{equation}
\end{subequations}
where $\mathcal{X} \subseteq \mathbb{R}^{n_x}$ is a compact region where the value function approximation is encouraged to be accurate.
The integral objective is to push up the under-estimate $\underline{J}$.
\subsection{Global value function over-approximation}
Similarly, a function $\bar J$ is a \textit{global} over-estimate of the value function \cite{jiang2015global} if it satisfies:
\begin{align}
    \forall x,\; \min_{u \in \mathcal{U}} \biggl[l(x, u) + \frac{\partial \bar{J}}{\partial x} f(x, u)\biggr]\le 0.\label{eq:HJB_ineq_ub}
\end{align}
The over-estimate HJB inequality is equivalent to 
\begin{align}
    \forall x, \exists u \in \mathcal{U}, \; l(x, u) + \frac{\partial \bar{J}}{\partial x} f(x, u) \le 0. 
    \label{eq: J_upper_inequality existential}
\end{align}
Handling the $\exists$ quantifier is much more challenging in general. However, for a fixed control law $u = \pi(x)\in\mathcal{U}$ that globally asymptotically stabilizes $x_d$, a sufficient condition that $\bar{J}$
 is an over-approximator is
 \begin{align}
    \forall x, \pi(x) \in \mathcal{U}, \; l(x, \pi(x)) + \frac{\partial\bar{J}}{\partial x}f(x, \pi(x))\le 0.
    \label{eq:J_upper_fixed_control}
\end{align}
A good over-approximator  satisfying \eqref{eq: J_upper_inequality existential} can be found in the manner of policy evaluation using the optimization program:
\begin{subequations} 
\label{eq:J_upper_fixed_control_program}
\begin{equation}
\label{eq:J_upper_fixed_control_program_a}
    \min \int_{\mathcal{X}} \bar{J}(x)\; dx 
\end{equation}
\vspace*{-0.1cm}
\begin{equation} 
\label{eq:J_upper_fixed_control_program_b}
    \text{s.t. } \; 
    l(x, \pi(x)) + \frac{\partial \bar{J}}{\partial x} f(x, \pi(x)) \leq 0, \quad  \forall \pi(x) \in \mathcal{U}.
\end{equation}
\end{subequations}
Similar to the under-estimate program \eqref{eq:J_lower_program}, the integral of $\bar J(x)$ over $\mathcal{X}$ is used to push down the over-estimate.

It is often impractical to find tight \textit{global} value function under-/over- approximators. In this paper, we restrict to a \textit{local} compact region $\mathcal{X}^{\text{h}} \subset \mathbb{R}^{n_x}$ ($h$ stands for HJB) where low-degree polynomials provide tight approximations and synthesize approximate optimal controllers that generate dynamic behavior for complicated robotic systems.

\section{METHOD}
The programs \eqref{eq:J_lower_program} and \eqref{eq:J_upper_fixed_control_program} search over the space of all functions and are therefore infinite dimensional. In this section, we discuss strong, finite-dimensional formulations of convex restrictions to \eqref{eq:J_lower_program} and \eqref{eq:J_upper_fixed_control_program} which can find good local under- and over-approximations of the value function over a compact domain using SOS, a convex optimization technique that searches over a parameterized family of polynomials through semidefinite programming\cite{parrilo2000structured}. We then discuss how to formulate the programs for rigid-body systems.
\subsection{Regions of interest}
When working with compact subsets of a control system, the notion of an \emph{invariant set} of the closed-loop system is important. 

\begin{definition}
Given a controller $\pi$ and the dynamical system $\dot{x} = f(x, \pi(x))$, a set $\mathcal{I}_{\pi}$ is an \emph{invariant set} if
 \begin{align*}
    x(0) \in \mathcal{I}_{\pi} \Rightarrow \; x(t) \in \mathcal{I}_{\pi} ,\forall~ t \in [0, \infty).
\end{align*}
The largest invariant set is the union of all invariant sets.
\end{definition}


We now define several regions which will be crucial for formulating the optimization programs to search for value function approximators. Their relationship is illustrated in Fig. \ref{fig:regions}:
\begin{itemize}
    \item $\mathcal{X}^{\text{h}}$: the region over which the HJB inequalities \eqref{eq: J_lower_inequality},\eqref{eq:J_upper_fixed_control} are enforced. 
    \item $\mathcal{B}_{\pi}^{\text{h}}$: the largest invariant set of the closed-loop system driven by a controller $\pi$ that is contained within $\mathcal{X}^{\text{h}}$.
    \item $\mathcal{X}$: the region over which we are interested in approximating the value function. This region must be chosen such that $\mathcal{X} \subseteq \mathcal{B}_{\pi^*}^{\text{h}} \subseteq \mathcal{X}^{\text{h}}$.
    \item $\mathcal{B}_{\pi}$: the largest invariant set of the closed-loop system driven by a controller $\pi$ that is contained within $\mathcal{X}$. 
\end{itemize}

\vspace*{-0.1cm}
\subsection{Local value function under-approximation} 
\label{subsection:lower_bound_method}
We parametrize $\underline{J}$ as a polynomial and require the HJB inequality \eqref{eq: J_lower_inequality} for the under-estimate to hold on the compact region $\mathcal{X}^{\text{h}}$. With polynomial $\underline{J}$, the nonnegativity constraint \eqref{eq: J_lower_inequality} can be enforced by SOS. 
We can then synthesize a tight under-approximator via the polynomial optimization program
\begin{align} \label{eq:J_lower_sos}
\underline{J}^* \coloneqq \argmax_{\underline{J}(x)\succeq 0}  \eqref{eq:J_lower_program_a} \text{ s.t. }  \eqref{eq:J_lower_program_b} \; \text{for } x \in \mathcal{X}^{\text{h}}, 
\end{align}
where $\underline{J}(x)\succeq 0$ is defined as $\underline J(x) \geq 0$ for all $x \in \mathcal{X}^{\text{h}}$ and $\underline J(x_{d}) = 0$. This positive-semidefinite requirement on $\underline J(x)$ is added because we know that the value function satisfies this condition, and the resulting $\underline J^*$ might be able to act as a Lyapunov function to certify the region of attraction for the closed-loop system using program \eqref{eq:roa}. One can easily cast \eqref{eq:J_lower_sos} (and the subsequent nonnegativity constraints for $x$ in a semialgebraic set) as a SOS program as shown in Appendix \ref{subsection:sos_program}. An approximate optimal controller can be generated from $\underline J^*(x)$: 
\begin{align}
    \underline{\pi}(x) = \Tilde \pi(x; \underline{J}^*)
\end{align}

\begin{prop} \label{prop:lower_bound}
The function $\underline J^*(x)$ is an under-estimate of the value function for $x \in \mathcal{B}_{\pi^*}^{\text{h}}$.
\end{prop}
The proof of Proposition \ref{prop:lower_bound} is in Appendix \ref{subsection:proposition1_proof}.

\subsection{Local value function over-approximation}
\label{subsection:upper_bound_method}
We similarly seek to enforce the HJB inequality \eqref{eq:J_upper_fixed_control} for the over-approximation on $\mathcal{X}^{\text{h}}$. Provided a polynomial control law $\pi_0(x)$ that asymptotically stabilizes $x \in \mathcal{X}^{h}$ to $x_d$, we can plug in $\pi_0$ as $\pi$ to search for a good over-estimate using the optimization program :
\begin{equation}
\begin{aligned}
\label{eq:J_upper_fixed_control_sos}
    \bar{J}^* \coloneqq \argmin_{\bar{J}(x)\succeq 0} \eqref{eq:J_upper_fixed_control_program_a} \text{ s.t. } \eqref{eq:J_upper_fixed_control_program_b} \; \text{for } x \in \mathcal{X}^{\text{h}}.
\end{aligned}
\end{equation}
An approximate optimal controller derived from $\bar J^*(x)$ is
\vspace*{-0.15cm}
\begin{align}
    \bar{\pi}(x) = \Tilde{\pi}(x; \bar{J}^*)
\end{align}
\begin{prop} \label{prop:upper_bound}
The function $\bar J^*(x)$ is an over-estimate of the value function for 
$x \in (\mathcal{B}_{\pi_0}^{\text{h}} \cup \mathcal{B}_{\bar \pi}^{\text{h}})\cap\mathcal{B}_{\pi^*}^{\text{h}}$.
\end{prop}
\begin{mproof}
Starting from $x(0) \in \mathcal{B}_{\pi_0}^{\text{h}}$, integrate the inequality \eqref{eq:J_upper_fixed_control} along the trajectory $(x,  \pi_0(x))$
\begin{align*}
    0 &\ge \int_0^{\infty} [l(x,  \pi_0(x)) + \dot{\bar{J}}^*(x)] \; dt \\
    &= \int_0^{\infty} l(x,  \pi_0(x)) \; dt + \bar{J}^*(x(\infty)) - \bar{J}^*(x(0))\\
    \bar{J}^*(x(0)) &\ge \int_0^{\infty} l(x, \pi_0(x)) \; dt = J^{ \pi_0} (x(0)),
\end{align*}
where the stabilizing controller $\pi_0$ leads to $x(\infty) = x_d$ and $\bar{J}^*(x(\infty))=0$. Therefore, we have $\forall x \in \mathcal{B}_{\pi_0}^{\text{h}}, J^{\pi_0}(x) \leq \bar J^*(x)$. Moreover, since $\bar \pi$ is the minimizer of \eqref{eq:HJB_ineq_ub}, we have that $\forall x \in \mathcal{X}^{\text{h}}, l(x, \bar \pi(x)) + \frac{\partial \bar{J}^*}{\partial x} f(x, \bar \pi(x)) \le 0$. We can then prove similarly $\forall x \in \mathcal{B}_{\bar \pi}^{\text{h}}, J^{\bar \pi}(x) \leq \bar J^*(x)$. Note that $\mathcal{B}_{\pi_0}^{\text{h}} \cup \mathcal{B}_{\bar \pi}^{\text{h}}$ is a control invariant set \cite{blanchini1999set}.

$\forall x \in \mathcal{B}_{\pi^*}^{\text{h}}, J^*(x) \leq J^{\pi_0}(x), J^*(x) \leq J^{\bar \pi}(x)$ since by definition, the value function leads to the optimal policy and thus incurs smaller cost than any other policies.
\end{mproof}
If $\pi_0(x) = \text{clamp}(\pi_{\text{poly}}(x), u_{\text{min}}, u_{\text{max}})$ is the saturation of a polynomial control law $\pi_{\text{poly}}(x)$, a piecewise analysis following the approach proposed in \cite{tedrake2010lqr} should be performed to impose input limits for the over-approximation. In particular, we enforce the following conditions on the polynomial optimization program \eqref{eq:J_upper_fixed_control_sos}:
\begin{equation} 
\label{eq:actuator_saturation_1}
\begin{aligned}
&\pi_{\text{poly}}(x) \geq u_{\text{max}}, x \in \mathcal{X}^{\text{h}} \Rightarrow l(x, u_{\text{max}}) + \frac{\partial \bar{J}}{\partial x} f(x, u_{\text{max}}) \leq 0 \\
&\pi_{\text{poly}}(x) \leq u_{\text{min}}, x \in \mathcal{X}^{\text{h}} \Rightarrow l(x, u_{\text{min}}) + \frac{\partial \bar{J}}{\partial x} f(x, u_{\text{min}}) \leq 0 \\
&u_{\text{min}} \leq \pi_{\text{poly}}(x) \leq u_{\text{max}}, x \in \mathcal{X}^{\text{h}} \Rightarrow \\
& \hspace*{2cm} l(x, \pi_{\text{poly}}(x)) + \frac{\partial \bar{J}}{\partial x} f(x, \pi_{\text{poly}}(x)) \leq 0, 
\end{aligned}
\end{equation}
which can be incorporated as SOS conditions with additional multipliers.
\begin{remark} 
A piecewise-polynomial control law can be obtained from a clamped LQR controller around the desired set point or the under-approximation controller $\underline{\pi}(x)$. The stabilizing region of the initial controller affects the maximum allowable size of $\mathcal{X}^{\text{h}}$.
\end{remark}
\begin{remark}
$\bar J^*$ is a control Lyapunov function \cite{sontag1983lyapunov} on $\mathcal{B}_{\pi_0}^{\text{h}} \cup \mathcal{B}_{\bar \pi}^{\text{h}} $ since 
\begin{align*}
\frac{\partial \bar{J}^*}{\partial x} f(x,  \pi_0(x))\le -l(x,  \pi_0(x)) < 0, \; \forall x \in \mathcal{B}_{\pi_0}^{\text{h}}\backslash\{x_d\} \\
    \frac{\partial \bar{J}^*}{\partial x} f(x, \bar \pi(x))\le -l(x, \bar \pi(x)) < 0, \; \forall x \in \mathcal{B}_{\bar \pi}^{\text{h}}\backslash\{x_d\}. 
\end{align*}
\end{remark}
\subsection{Formulation - Rigid-body systems}
\label{subsection:rigid_body_system}
The dynamics of rigid-body systems can be written as polynomial functions through a change of state variables, see \cite{posa2013lyapunov, shen2020sampling} for a complete description. Here we use the simple pendulum as an example.

\begin{example}
By denoting the new coordinate variable ${s\equiv\sin\theta,}~ {c\equiv\cos\theta}$, the simple pendulum dynamics given in \cite{tedrake2009underactuated}[Chapter 2, (1)] with mass $m$, length $l$, damping ratio $b$, and gravity $g$ can be converted to polynomial dynamics 
\begin{align}
    x = \begin{bmatrix}s\\c\\\dot \theta \end{bmatrix}, \; \dot x = \begin{bmatrix}c \dot \theta\\-s \dot \theta\\-\frac{1}{ml^2} (b \dot \theta + mgls)\end{bmatrix} + \begin{bmatrix}0\\0\\\frac{1}{ml^2}\end{bmatrix}u.
\end{align}
with the compact domain of $x$ being $\mathcal{X}^{\text{h}} = \{x \in \mathbb{R}^{n_x}| s^2 + c^2=1, \dot \theta_{\min}^{\text{h}} \leq \dot \theta \leq \dot \theta_{\max}^{\text{h}}\}$ and $\mathcal{X} = \{x \in \mathbb{R}^{n_x}| s^2 + c^2=1, \dot{\theta}_{\min} \leq \dot \theta \leq \dot{\theta}_{\max}\}$. The integration in  \eqref{eq:J_lower_program_a} can be written as
\begin{align*}
\int_{\dot{\theta}_{\min}}^{\dot{\theta}_{\max}} \int_{s^2 + c^2=1} \underline{J}(s, c, \dot \theta)\; dx.
\end{align*}
\end{example}

\section{Regional analysis}
\label{section:regional_analysis}
\begin{figure}
\centering
	\includegraphics[width=0.48\textwidth]{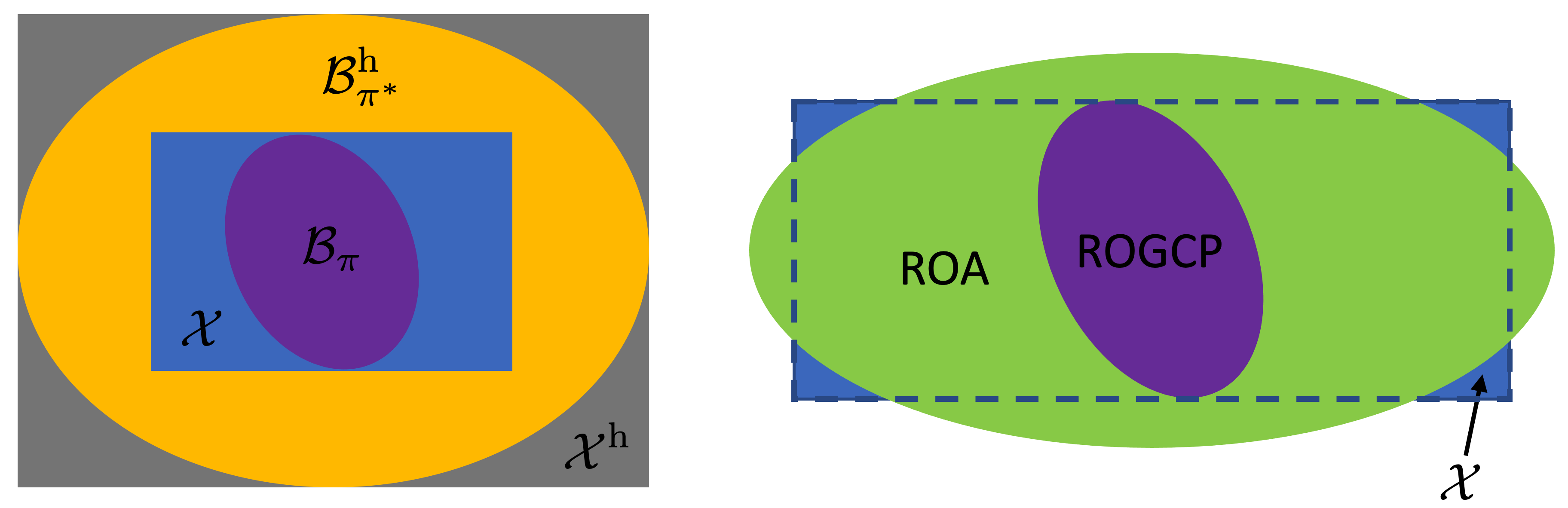}
	\caption{Visualization of different regions. $\mathcal{X}^{\text{h}}$ is the region where the HJB inequalities are satisfied. $\mathcal{B}_{\pi^*}^{\text{h}}$ is the largest invariant set of the closed-loop system driven by $\pi^*$ contained within $\mathcal{X}^{\text{h}}$. $\mathcal{X}$ is the region where the value function approximation is encouraged to be accurate. ROGCP illustrates the invariant region where the synthesized controllers have stability and suboptimality guarantees. ROGCP is strictly contained within $\mathcal{X}$ and lies within the closed-loop system's ROA. In general, the closed loop system's ROA and $\mathcal{X}$ have no containment relationship.}  
	\label{fig:regions}
	\vspace*{-0.5cm}
\end{figure}
In this section, we introduce various regions over which we evaluate the quality of our value function approximations and the resulting synthesized controllers. Fig. \ref{fig:regions} illustrates an example of the containment relationship between these various sets whose defining properties are summarized in Table \ref{table:regions}.
\subsection{Region of guaranteed controller performance}
Although our value function approximations satisfy the respective HJB inequalities on $\mathcal{X}^{\text{h}}$, the corresponding controllers might not be able to certifiably stabilize the entire region since $\mathcal{X}^{\text{h}}$ is not necessarily an invariant set for the resulting closed-loop system. In addition, the value function over-approximation is on the set $(\mathcal{B}_{\pi_0}^{\text{h}} \cup \mathcal{B}_{\bar \pi}^{\text{h}})\cap\mathcal{B}_{\pi^*}^{\text{h}}$ while the under-approximation is on $\mathcal{B}_{\pi^*}^{\text{h}}$. Unfortunately, the invariant set $\mathcal{B}_{\pi^*}^{\text{h}}$ is difficult to compute exactly. However, since $\mathcal{X} \subseteq \mathcal{B}_{\pi^*}^{\text{h}}$, we can restrict our calculation to $\mathcal{X}$ and try to find good inner approximations of the largest invariant set $\mathcal{B}_{\bar \pi} \subseteq \mathcal{X}$ (resp. $\mathcal{B}_{\underline \pi}$) where the HJB inequality \eqref{eq:J_upper_fixed_control}  (resp. \eqref{eq: J_lower_inequality}) is always satisfied along the trajectories driven by the synthesized controller $\bar \pi$ (resp. $\underline \pi$).


\begin{definition} The region of guaranteed controller performance (ROGCP) for the closed-loop dynamics under $\bar \pi$ (resp. $\underline \pi$) is defined as the largest invariant set contained in $\mathcal{X}$.
\end{definition}
\subsubsection{Approximation of $\bar \pi$'s ROGCP}
The sublevel set of a Lyapunov function is commonly used to characterize an invariant set.
We can search for an inner approximation of the largest invariant set strictly contained within the compact region $\mathcal{X}$ using a $\bar{\rho}$-sublevel set of $\bar J^*$:
\begin{gather}
\label{eq:region_of_approximation}
    \bar \rho \coloneqq \min_{x \in \partial \mathcal{X}} \bar J^*(x)\\
    \hat{\mathcal{B}}_{\bar \pi}^{\bar \rho} \coloneqq \{x \in \mathcal{X}|\bar J^*(x) < \bar \rho\},
\end{gather}
where $\partial \mathcal{X}$ is the boundary of $\mathcal{X}$. 

\begin{prop}\label{prop:upper_connected}
$\hat{\mathcal{B}}_{\bar \pi}^{\bar \rho}$ is a bounded, connected, invariant set.
\end{prop}
We prove Proposition \ref{prop:upper_connected} in Appendix \ref{subsection:proposition3_proof}. 

Within $\bar \pi$'s ROGCP, the following cost-to-go inequalities hold.
\begin{prop} \label{prop:upper_bound_ineq}
$\forall x \in \hat{\mathcal{B}}_{\bar \pi}^{\bar \rho}, \; J^*(x) \leq J^{\bar \pi}(x) \le \bar J^*(x).$
\end{prop}
The proof of Proposition \ref{prop:upper_bound_ineq} is similar to Proposition \ref{prop:upper_bound}.
\subsubsection{Approximation of $\underline \pi$'s ROGCP}
Unlike $\bar{J}^*$, $\underline J^*$ is not a Lyapunov function by construction and its sublevel set $$\hat{\mathcal{B}}_{\underline \pi}^{\rho} \coloneqq \{x \in \mathcal{X}|\underline J^*(x) < \rho \}$$ is not guaranteed to be invariant. Nonetheless, we can search for the region where $\underline J^*$ acts as a Lyapunov function by enforcing the condition that
\begin{equation*}\footnotesize
\begin{aligned}
    \frac{\partial \underline{J}^*}{\partial x} f(x, \underline \pi(x)) \le - \epsilon (m(x)-m(x_d))^T(m(x)-m(x_d)) \text{ for } x \in \hat{\mathcal{B}}_{\underline \pi}^{\rho},
\end{aligned}
\end{equation*}
where $\epsilon$ is a small positive constant and $m(x)$ is a vector of monomials in $x$. The largest $\hat{\mathcal{B}}_{\underline \pi}^{\rho}$ can be obtained with bisection search on $\rho$ in the optimization program:
\begin{align}
\label{eq:rohjb_lower}
\begin{gathered}
    \underline \rho \coloneqq \max_{\rho, \lambda(x)} \rho \\ 
    \text{s.t. } \underline J^*(x) - \rho \ge 0 \text{ for } x \in \partial \mathcal{X}\\
      - \frac{\partial \underline{J}^*}{\partial x} f(x, \underline \pi(x)) - \epsilon (m(x)-m(x_d))^T(m(x)-m(x_d)) + \\ \lambda(x) (\underline J^*(x) - \rho) \ge 0 \text{ for } x \in \mathcal{X}, \underline \pi(x) \in \mathcal{U} \\
    \lambda(x) \ge 0 \text{ for } x \in \mathbb{R}^{n_x}.
\end{gathered}
\end{align}
For piecewise polynomial $\underline \pi$, we can perform the piecewise analysis similar to \eqref{eq:actuator_saturation_1} on program \eqref{eq:rohjb_lower}.

The following propositions are the direct analogues of Proposition \ref{prop:upper_connected} and \ref{prop:upper_bound_ineq} for $\hat{\mathcal{B}}_{\underline \pi}^{\underline \rho}, \underline J^*$ and $J^{\underline \pi}$.

\begin{prop}\label{prop:lower_connected}
$\hat{\mathcal{B}}_{\underline \pi}^{\underline \rho}$ is a bounded, connected, invariant set.
\end{prop}
\begin{prop} \label{prop:lower_bound_approx}
$\forall x \in \hat{\mathcal{B}}_{\underline \pi}^{\underline \rho}, \; \underline J^*(x) \le J^*(x) \leq J^{\underline \pi}(x) .$
\end{prop}
\begin{table}
    \centering
    \def\arraystretch{1.3}
    \begin{tabular}{ |c|c|c| } 
    \hline
      & HJB inequality satisfied & Invariant \\
    \hline
    $\mathcal{X}$ & yes & not guaranteed \\ 
    \hline
    ROGCP & yes & yes\\ \hline
    ROA & not guaranteed & yes \\ \hline
    \end{tabular}
    \caption{HJB inequality satisfaction and set invariance of different regions.}
    \label{table:regions}
    \vspace*{-0.8cm}
\end{table}

\subsection{Region of attraction}
Notice that $\bar{J}^*$ satisfies the Lyapunov condition on $\mathcal{X}^{\text{h}}$, that is $\dot{\bar J}^*(x) < 0$ for all $x \in \mathcal{X}^{\text{h}} \backslash\{x_d\}$. We can find an inner approximation of the region of attraction (ROA) for the closed-loop system driven by $\bar \pi$ and verified by $\bar{J}^*$ as a Lyapunov function using the optimization program studied in \cite{shen2020sampling}:
\begin{align}
\label{eq:roa}
    \begin{gathered}
    \rho^* 
    \coloneqq \max_{\rho, \lambda(x)} \rho\\
    \text{s.t. } \; ((x-x_d)^T(x-x_d))^d (\bar{J}^*(x) - \rho) \\
    + \lambda(x)\frac{\partial \bar{J}^*}{\partial x} f(x, \bar \pi(x)) \geq 0 \text{ for } \bar \pi(x) \in \mathcal{U},
    \end{gathered}
\end{align}
where $d$ is a positive integer and $\lambda(x)$ is a free polynomial. 
\begin{remark}
Define the sublevel set $\mathcal{C}_{\rho^*} \coloneqq \{x|\bar J^*(x) < \rho^*\}$. Note that the optimization program \eqref{eq:roa} ensures that $\dot{\bar J}^*(x) < 0, \forall x \in \mathcal{C}_{\rho^*}\backslash\{x_d\}$. Therefore, the connected component of $\mathcal{C}_{\rho^*}$ that includes the origin is an invariant set and an inner approximation of the true region of attraction for the closed-loop system under $\bar \pi$. 
\end{remark}

\section{EXPERIMENTAL RESULTS}
\label{section:results}
We synthesize value function approximations and their corresponding controllers for the inverted pendulum, cart-pole, and quadrotor as examples of continuous-time, rigid-body systems. A value function under-estimate and its corresponding controller are obtained for a planar pushing task with hybrid dynamics. All programs are run on an Intel Core i9-7900X CPU and solved using Mosek. Table \ref{table:experiments} records the function degree and computation time to synthesize a controller that stabilizes a large region of the state space for each task.

\begin{table}
    \centering
    \def\arraystretch{1.3}
    \begin{tabular}{ |c|c|c|c|c|c| } 
    \hline
      & $n_x$ & $\underline{J}^*$ deg &$\underline{J}^*$ time (s) & $\bar J^*$ deg & $\bar J^*$ time (s)\\
    \hline
    pendulum & 3 & 2 & 0.044 & 2 & 0.286 \\ 
    \hline
    cart-pole & 5 & 6 & 540 & 2 & 6.4\\ \hline
    quadrotor & 13 & 2 & 30 & 2& 1100\\ \hline
    pusher & 6 & 2 & 1410 & --- & --- \\ \hline
    \end{tabular}
    \caption{Degree and computation time of value function approximations to obtain a stabilizing controller for each task.}
    \label{table:experiments}
    \vspace*{-0.5cm}
\end{table}

\vspace{-0.2cm}
\subsection{Inverted pendulum}
We synthesize an under- and over-approximation to the value function and their corresponding stabilizing controllers of an inverted pendulum. We aim to swing up and stabilize the pendulum at the upright equilibrium $\theta = \pi, \dot \theta = 0$.
\begin{figure}
\centering
	\includegraphics[width=0.47\textwidth]{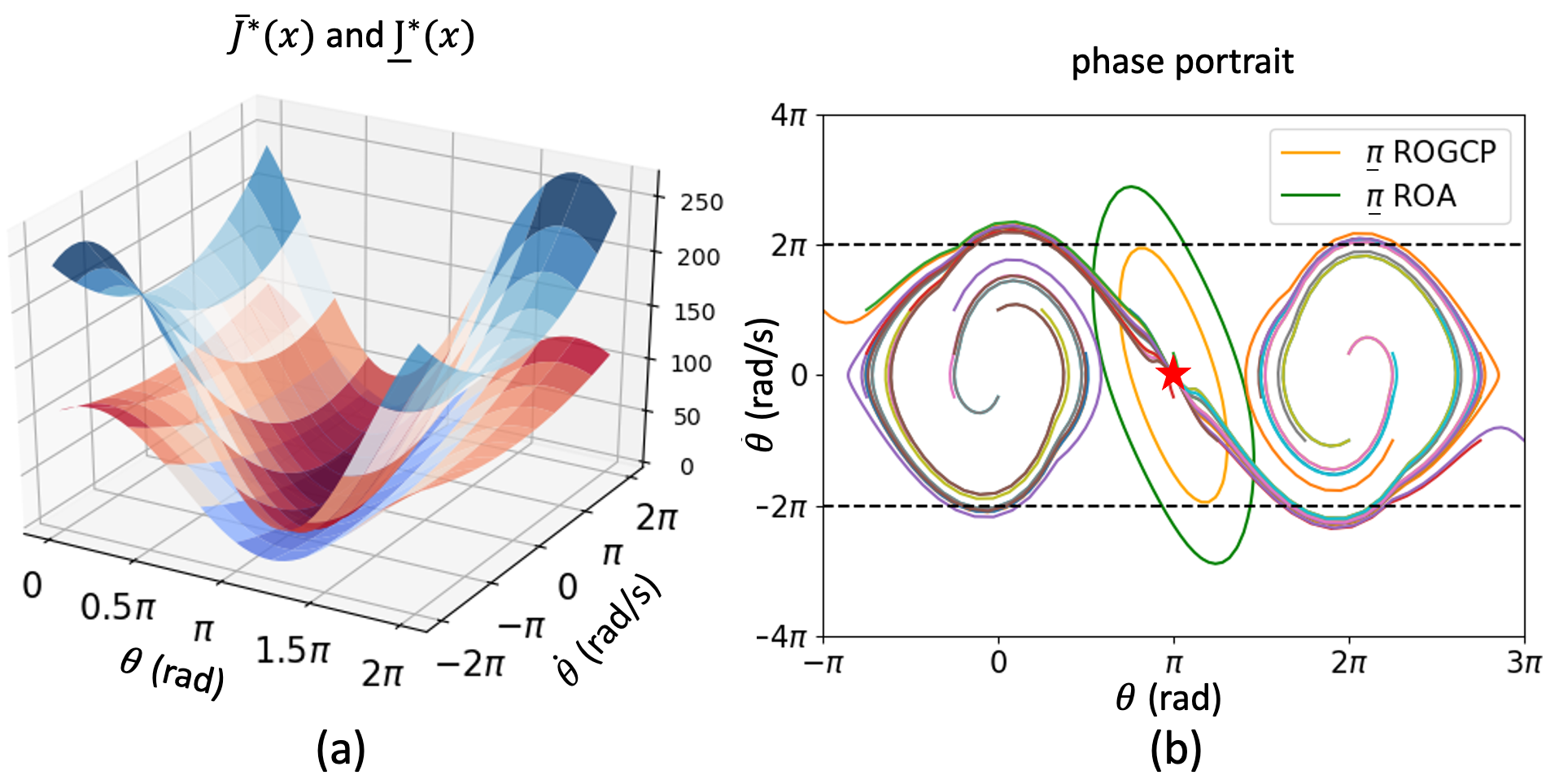}
	\caption{Inverted pendulum: (a) quadratic value function under- and over- estimate with unconstrained control input; (b) inner approximation of region of guaranteed controller performance/region of attraction and phase portrait of simulating the pendulum using the under-estimate controller with input limits. The orange contour is saturated $\underline \pi(x)$'s ROGCP and the green curve displays the closed-loop system's ROA driven by $\underline \pi(x)$. The dashed black line is the region of interest $\mathcal{X}$.}
	\label{fig:pendulum}
	\vspace*{-0.5cm}
\end{figure}
The region of interest $\mathcal{X}$ is $[s,c,\dot{\theta}] \in [\pm 1, \pm 1, \pm 2\pi]$ and the HJB inequalities are verified in the region $[s,c,\dot{\theta}] \in [\pm 1, \pm 1, \pm 3\pi]$ as $\mathcal{X}^{\text{h}}$. A quadratic value function under-estimate (over-estimate)  $\underline J^*(x)$ ($\bar{J}^*(x)$) with unconstrained control input is visualized in Fig. \ref{fig:pendulum}a. 
The phase portrait in Fig. \ref{fig:pendulum}b demonstrates that starting from any initial state, the system can always converge to the upright equilibrium with input limits $1.8\ \mathrm{N \cdot m}$
(the gravity torque is $mgl = 4.9\ \mathrm{N \cdot m}$). The spiraling trajectories demonstrate that the controller performs nontrivial pumping to swing up the pendulum. 

We can search for the inner approximation of ROA where $\underline{J}^*(x)$ serves as a Lyapunov function using the polynomial optimization program \eqref{eq:roa} by replacing $\bar J^*(x)$ and $\bar \pi(x)$ with $\underline{J}^*(x)$ and $\underline \pi(x)$. In Fig. \ref{fig:pendulum}b, the orange contour displays saturated $\underline \pi$'s ROGCP and the green curve demonstrates the inner approximation of the closed-loop system  ROA, driven by the under-estimate controller and verified by $\underline J^*(x)$.

\subsection{Cart-pole}
\begin{figure*}[t]
\centering
	\includegraphics[width=1\textwidth]{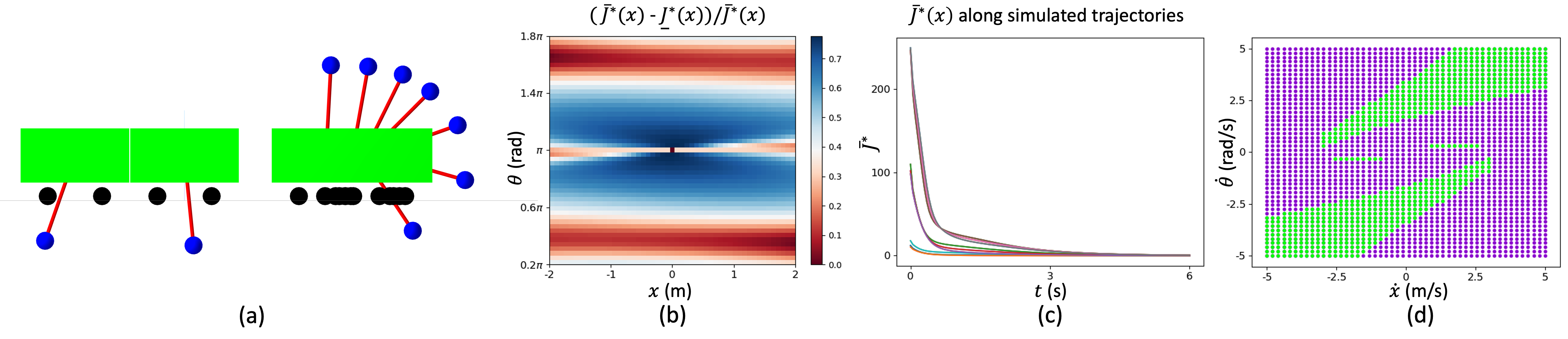}
	\caption{Cart-pole: (a) snapshots of the full swing-up simulation using $\underline{\pi}(x)$, an animation can be found \href{https://hongkai-dai.github.io/figures/cartpole.html}{here}; (b) $(x, \theta)$ slice of the value function relative approximation error; (c) the value function over-estimate $\bar{J}^*(x)$ always decreases along the trajectories starting within $\bar{J}^*(x)$'s ROA and simulated with $\bar{\pi}(x)$; (d) $(\dot x, \dot \theta)$ slice of sampled initial states that can be stabilized by the SOS/LQR controller.} 
	\label{fig:cart-pole}
	\vspace*{-0.5cm}
\end{figure*}
We apply our method to swing up and balance a cart-pole with five states and one actuator \cite{tedrake2009underactuated} around the unstable equilibrium $\theta = \pi$. We are interested in the region $[x, s, c, \dot x, \dot \theta] \in [\pm2, \pm 1, \pm 1, \pm 5, \pm 5]$ as $\mathcal{X}$ and verify the HJB inequalities in the box region $[x, s, c, \dot x, \dot \theta] \in [\pm3, \pm 1, \pm 1, \pm 6, \pm 6]$ as $\mathcal{X}^{\text{h}}$. With input limits $100\ \mathrm{N \cdot m}$ (cart mass $10\ \mathrm{kg}$, pole mass $1\ \mathrm{kg}$, pole length $0.5\ \mathrm{m}$), a 6-degree value function under-estimate is required to swing up the cart-pole while a quadratic value function over-estimate with an LQR initial controller $\pi_0$ suffices to accomplish the task. The successful full swing-up of the cart-pole, accomplished with a nontrivial pumping, is visualized in Fig. \ref{fig:cart-pole}a. Fig. \ref{fig:cart-pole}b illustrates the relative approximation error between the value function over- and under- approximator. Fig. \ref{fig:cart-pole}c shows that $\bar J^*(x)$ always decreases along the simulated trajectories from various initial states within the ROA, corroborating that the synthesized over-estimate is a Lyapunov function. In Fig. \ref{fig:cart-pole}d, we sample 2500 initial states within the box region $[\dot x, \dot \theta] \in [\pm 5, \pm 5]$ with $[x, \theta] = [0, 0]$, and color them based on whether the SOS/LQR controller succeeds in stabilizing the initial state to the goal (we observe that our SOS controller can stabilize all the initial states in this 2-dimensional slice)
\begin{itemize}
    \item Purple: SOS succeeds, LQR fails
    \item Green: both SOS and LQR succeed.
\end{itemize}
\subsection{Quadrotor}
\begin{figure}
\centering
	\includegraphics[width=0.47\textwidth]{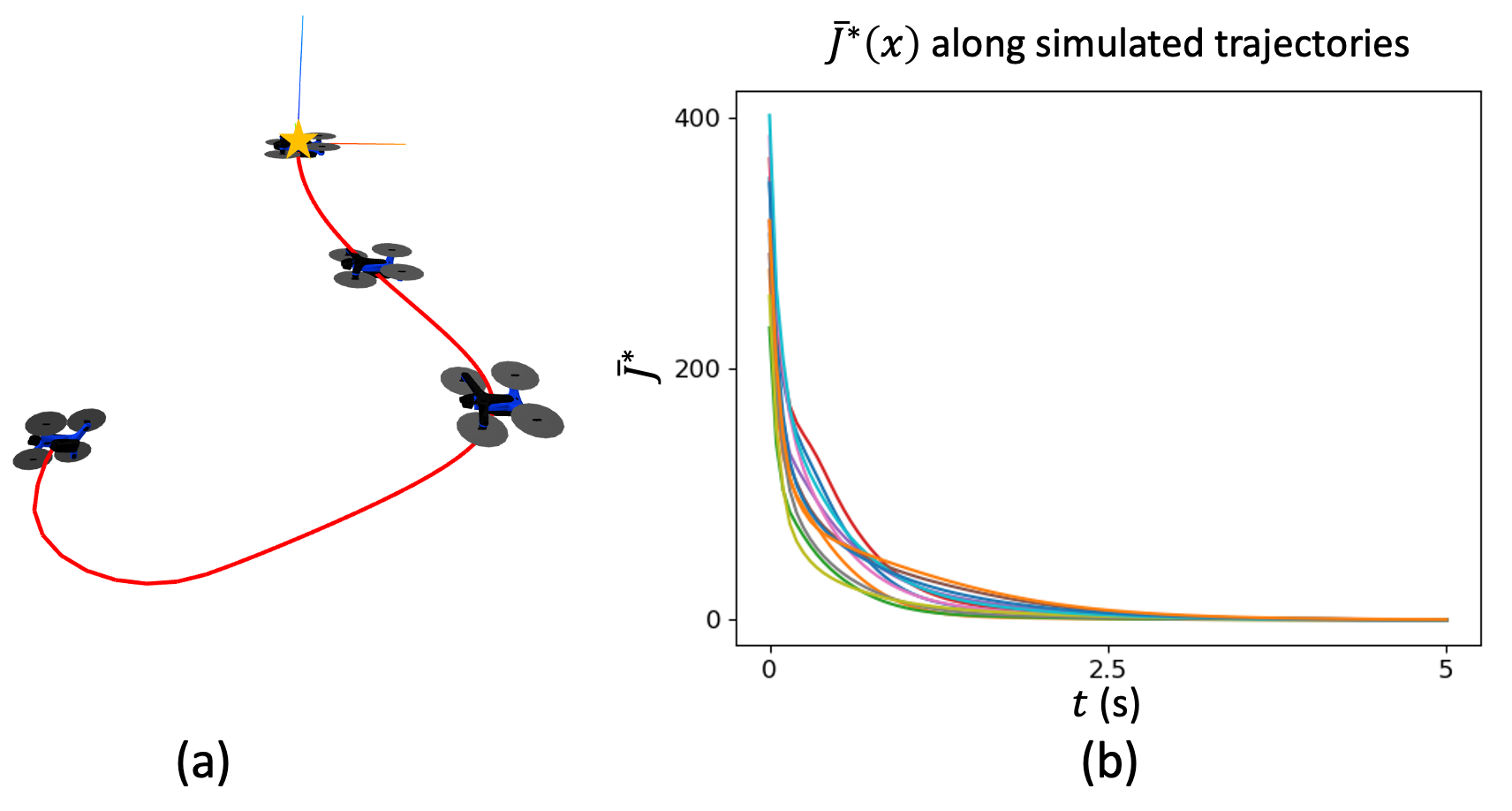}
	\caption{Quadrotor: (a) snapshots of the simulation using $\bar \pi(x)$. (b) The quadratic value function over-approximation $\bar J^*(x)$ along the trajectories starting within $\bar{J}^*(x)$'s ROA and simulated with $\bar{\pi}(x)$.} 
	\label{fig:quadrotor3d}
	\vspace*{-0.5cm}
\end{figure}
We test our proposed method on a quadrotor model with 13 states (quadrotor orientation is parameterized by a unit quaternion) and 4 actuators \cite{fresk2013full}. Our goal is to steer the quadrotor to hover at the origin. Let $\vb{x} = [x, y, z, \phi, \theta, \psi, \dot x, \dot y, \dot z, \omega_x, \omega_y, \omega_z]$ be the 12-dimensional state vector with the quadrotor's orientation parameterized by Euler angles, $\vb{x}^{\text{h}} = [1, 1, 1, 0.5\pi, 0.2 \pi, 0.5\pi, 1, 1, 1, 1, 1, 1]$. The HJB inequalities can be verified within the bounding box region $-\vb{x}^{\text{h}} \leq \vb{x} \leq \vb{x}^{\text{h}}$ as $\mathcal{X}^{\text{h}}$ with input constraints $2.5 \frac{mg}{4}$. As visualized in Fig. \ref{fig:quadrotor3d}a, the controller obtained from the value function over-estimate with actuator saturation can stabilize the quadrotor with initial conditions as large as $[1, 1, 1, \pi, 0.4 \pi, \pi, 8, 8, 8, 8, 8, 8]$.

Our method provides a tight quadratic under- and over- approximation on the true value function for the quadrotor system, as manifested by the small relative approximation error $\frac{\bar J^*(x) - \underline{J}^*(x)}{\bar J^*(x)}$ in Fig. \ref{fig:quadrotor3d_banner}b. We simulate the system starting from various initial conditions within the region of interest and observe that $\bar J^*(x)$ decreases along the simulated trajectories in Fig. \ref{fig:quadrotor3d}b.
\vspace*{-0.2cm}
\subsection{Planar pushing}
\label{subsection:planar_pushing}
\begin{figure}
\centering
	\includegraphics[width=0.2\textwidth]{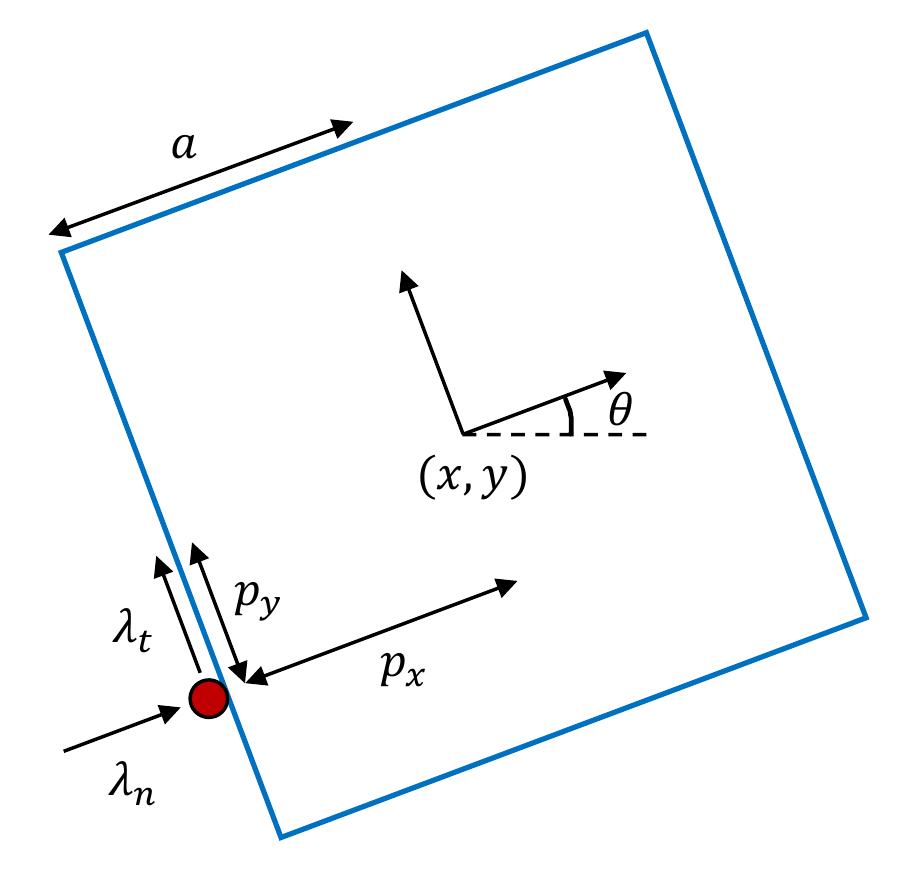}
	\caption{The planar pusher system with one point pusher and a 4-face slider.} 
	\label{fig:planar_pushing_system}
	\vspace*{-0.5cm}
\end{figure}
The under-approximation method can be extended to handle certain hybrid dynamical systems. Here, we use the planar pushing system (illustrated in Fig. \ref{fig:planar_pushing_system}) as an example. 

We adopt the quasi-static model \cite{trinkle2005time} for robotic manipulation systems. We model this as a hybrid dynamical system \cite{branicky1994unified} with 4 continuous modes (one mode for every slider face). The state is denoted as $x = [x_{\text{pusher}}, x_{\text{slider}}]$ with a mode index $x_{\text{mode}}\in\{1, 2, 3, 4\}$, where $x_{\text{pusher}}, x_{\text{slider}}$ are the pusher/slider state with continuous values. The control is denoted as $u = [u_{\text{mode}}, u_{\text{pusher}}]$ where $u_{\text{mode}}$ takes the discrete mode value, and $u_{\text{pusher}}$ takes the continuous value whose meaning changes based on $u_{\text{mode}}$ as we will see shortly.

At each time instant, the controller determines whether the pusher should stay on the current face or transition to a different face with $u_{\text{mode}}$. When the contact face remains unchanged ($u_{\text{mode}}$ remains unchanged), the control action includes the pusher velocity $u_{\text{pusher}} = v_{\text{pusher}}$. The continuous dynamics  (\cite{hogan2018reactive, bauza2018data}) are denoted as
\begin{align}
    & \dot{x}(t) = f_{x_{\text{mode}}}(x, u), \; x_{\text{mode}} = u_{\text{mode}}, \label{eq:mode_dynamics}
\end{align}
when $\lim_{\tau \rightarrow t^{-}} u_{\text{mode}}(\tau) = \lim_{\tau \rightarrow t^{+}} u_{\text{mode}}(\tau)$.
If the controller chooses to transition to a different mode (the pusher transitions to a different face) with $\lim_{\tau \rightarrow t^{-}} u_{\text{mode}}(\tau) \neq \lim_{\tau \rightarrow t^{+}} u_{\text{mode}}(\tau)$, the mode transition dynamics to post-transition state $x^+$ are
\begin{align}
\underbrace{x^+_{\text{pusher}} = u_{\text{pusher}},\;
x^+_{\text{slider}} = x_{\text{slider}}^{-}}_{x^+\coloneqq\Delta(x, u)},\; x^+_{\text{mode}} = u_{\text{mode}},\label{eq:controlled_jump}
\end{align}
namely, the controller chooses the contact face $x^+_{\text{mode}}$ and pusher position $x^+_{\text{pusher}}$ after face transition. Due to the quasi-static assumption, the slider remains static during the transition. We assume the transition is instantaneous, with a mode transition cost $l_m(x, u) = |u_{\text{pusher}} - x_{\text{pusher}}|^2>\varepsilon > 0$. 

Since each face transition cost $l_m(x, u)$ is strictly larger than an absolute constant, there will only be a finite number of face transitions in any trajectory with finite cumulative cost. The controller chooses to transition to a mode $k$ different from the current mode when the new state in mode $k$ has a smaller value of the optimal cost-to-go (or its under-estimate $\underline J^*$) than that of the current mode under the best continuous action. Such mode transitions are modeled as instantaneous events where the control input causes a discontinuity in the state. We denote the time of mode switch as $t_s, s=1,\hdots, N$. The optimal control problem is to minimize the cost function
\begin{equation}
\begin{aligned}
    \int^{\infty}_0 l(x, u) dt + \sum_{s=1}^N l_m(x(t^{-}_{s}), u(t_s)).
\end{aligned}
\end{equation}
It is shown in \cite{hedlund1999optimal} that $\underline J$ is a global under-estimate to the optimal cost-to-go for such hybrid systems if it satisfies \eqref{eq:HJB_ineq_lo} for all $f_k$ and:
\begin{equation}\footnotesize
\label{eq:hybrid_hjb_lower}
    \forall x(t^{-}_{s}), \min_{u} \biggl[l_m(x(t^{-}_{s}), u) + \underline J(\Delta(x(t_s^-), u)) - \underline J(x(t^{-}_{s}))\biggr] \geq 0.
\end{equation}
\normalsize

In addition, friction can be incorporated into our framework with the complementarity formulation \cite{posa2013lyapunov} 
\begin{subequations}
\label{eq:friction_constraints}
\begin{align}
v_t \lambda_t &\leq 0 \\
\mu^2 \lambda_n^2 - \lambda_t^2 &\geq 0\\
(\mu^2 \lambda_n^2 - \lambda_t^2)v_t &= 0,
\end{align}
\end{subequations}
where $v_t$ is the tangential velocity of the pusher, $\lambda_n, \lambda_t$ are the normal and tangential contact forces of the pusher on the slider and $\mu$ is the coefficient of friction. In the quasi-static setting, the velocity $v_t$ is part of the control $u$, and the normal and tangential forces $\lambda_n, \lambda_t$ are neither part of the state $x$ nor control $u$. By treating $\lambda_{n}$ and $\lambda_{t}$ as new indeterminates, we can define the set $\mathcal{D} = \{(u, \lambda_{n}, \lambda_{t}) \mid \text{satisfying \eqref{eq:friction_constraints}}\}$.

Similar to Sec. \ref{subsection:lower_bound_method}, we can enforce the HJB inequality \eqref{eq:hybrid_hjb_lower} on the compact region $\mathcal{X}^{\text{h}}$. We can extend the under-approximation program \eqref{eq:J_lower_sos} to the hybrid system case with friction via the program
\begin{equation}
\begin{aligned} \small
\label{eq: J_lower_bound_contact}
\begin{gathered}
    \underline J^* \coloneqq \argmax_{\underline{J}(x) \succeq 0} \int_{\mathcal{X}} \underline{J}(x)\; dx\\
    \text{s.t. } \; \text{for all } k: \\ 
    l(x, u) + \frac{\partial \underline{J}}{\partial x} f_k(x, u) \geq 0 \; \text{for } x \in \mathcal{X}^{\text{h}}, u \in \mathcal{U}, (u,\lambda_n, \lambda_t)\in \mathcal{D} \\
    l_m(x, u) + \underline J(\Delta(x, u)) - \underline J(x) \geq 0 \text{ for }x, \Delta(x, u) \in \mathcal{X}^{\text{h}}, u \in \mathcal{U}.
\end{gathered}
\end{aligned}
\end{equation}

The complementarity constraints on the input preclude an analytic form of the control input as in \eqref{eq: minimizing_controller_closed_form}. Instead, a control input $\underline \pi$ at a particular state $x$ can be synthesized by solving the nonlinear program for the continuous dynamics:
\begin{align}
    \min_{u \in \mathcal{U}, \lambda_n, \lambda_t} \biggl[l(x, u) + \frac{\partial \underline{J}^*}{\partial x}f_k(x, u)\biggr] \\
    \text{s.t. } (u, \lambda_n, \lambda_t)\in\mathcal{D}. \nonumber
    \label{eq:contact_continuous_controller}
\end{align}
Similarly, the optimal action of the mode transition dynamics at a state $x$ can be found with the mixed-integer program:
\begin{equation}
\label{eq: mode_switch_mip}
\begin{gathered}
 \min_{u \in \mathcal{U}} \biggl[l_m(x, u) + \underline J^*(\Delta(x, u)) - \underline J^*(x)\biggr] \\
    \text{s.t. } \; \Delta(x, u)\in \mathcal{X}_k \text{ iff } b_k = 1.\; b_k \in \{0, 1\}, \textstyle\sum_k b_k = 1,
\end{gathered}
\end{equation}
where $b_k$ is the binary variable indicating if $\Delta(x, u)$ is in mode $k$ and $\mathcal{X}_k \subset \mathcal{X}^{\text{h}}$ is the compact set of mode $k$. During online execution, we calculate the optimal continuous control $\underline \pi(x)$ and mode switch $\Delta(x, u)$, and choose the control that incurs the smaller cost-to-go.

\begin{figure}
\centering
	\includegraphics[width=0.45\textwidth]{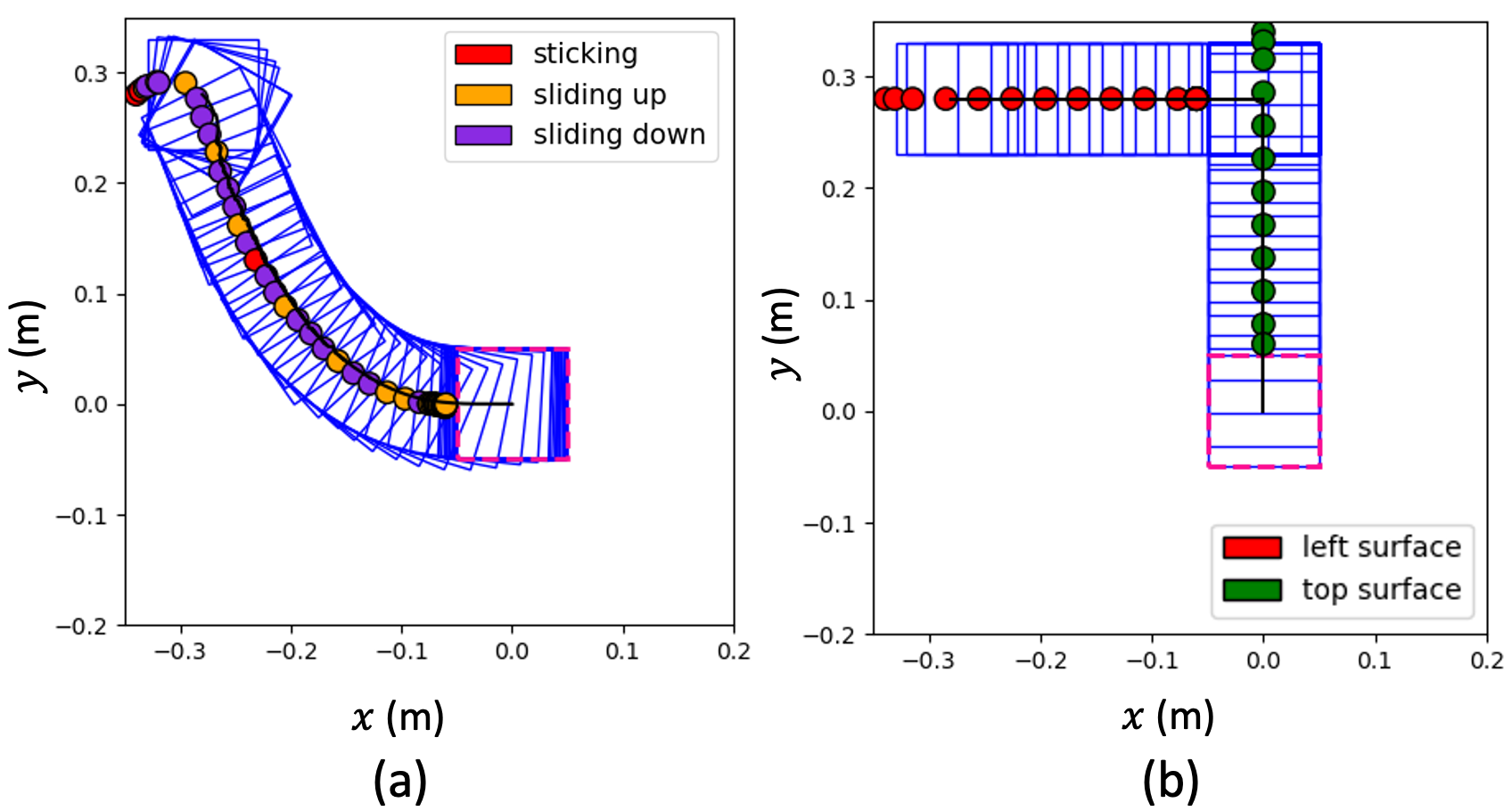}
	\vspace*{-2mm}
	\caption{Planar pushing: (a) snapshots of the simulation using $\underline{J}^*(x)$ starting from $[-0.28, 0.28, 0]$. The black curve is the slider's center of mass trajectory, the circle is the finger pusher and the pink dashed square is the desired slider pose. Different colors show that the finger pusher switches among 3 modes: red, orange and purple denote the finger sticking, sliding up and down along the slider's left face. (b) snapshots of the simulation involving face change using $\underline{J}^*(x)$. After finishing pushing the slider to the right, the finger pusher decides to transition to the top face of the slider (green circle denotes the face change) and then push it towards the origin.} 
	\label{fig:planar_pushing}
	\vspace*{-0.5cm}
\end{figure}

We synthesize a quadratic value function under-estimate and a controller with actuator saturation for the planar pushing task. The compact region
\begin{align*}
\mathcal{X} = &\{x|(p_x+a)(p_x-a)(p_y+a)(p_y-a)=0, \\ & -a \leq p_x \leq a, -a \leq p_y \leq a\},     
\end{align*}
\normalsize
requires the pusher to stay on the slider's surface, where $p_x, p_y$ are the slider's position and $a$ is half of the slider's side length. In Fig. \ref{fig:planar_pushing}, we present two trajectories of the circle pusher pushing the blue slider from $[x, y, \theta] = [-0.28, 0.28, 0]$ to the origin. For Fig. \ref{fig:planar_pushing}a, the pusher is restricted to be in contact with the slider's left surface while it is allowed to transition to other faces in Fig. \ref{fig:planar_pushing}b. This is a nontrivial task even for humans, and the mode switches include the finger sticking, sliding up, and sliding down on the slider's left surface (indicated by red, orange and purple circles). Fig. \ref{fig:planar_pushing}b shows a trajectory that involves the finger changing to a different face of the slider in order to push the slider towards the origin. The finger starts from the left surface and pushes the slider to the position right above the origin. Then the finger pusher decides to transition to the top surface (green circle indicates the finger's face change) and push the slider downward.
\subsection{Computation}
The computation time of the SOS programs depends both on the system dimension $n_x$ and the degree of the value function approximation. Table \ref{table:experiments} outlines the function degrees and computation time to synthesize the value function approximations with stabilizing controllers for each task. In general, $\underline J^*(x)$ takes less time to synthesize than $\bar J^*(x)$ of the same degree because the fixed control law $\pi(x)$ raises the final over-estimate SOS program to higher degrees. However, because the under-estimate is not guaranteed to be a Lyapunov function, a low-degree $\underline J^*(x)$ might not be able to stabilize a large region in the state space. For instance, the cart-pole needs a 6th-degree $\underline J^*(x)$ to swing up the system and stabilize the state space, resulting in 9-minute computation compared to 6.4 seconds for a quadratic $\bar J^*(x)$.
\section{CONCLUSIONS}
Building upon prior works on SOS hierarchy, we have summarized a unified approach to synthesizing regional under- and over- approximations to the value function via sums-of-squares programming, and synthesize dynamic controllers for various underactuated robotic systems. We performed local analysis on these value function approximations by computing the region of guaranteed controller performance and an inner approximation of the region of attraction using additional SOS programs. Moreover, we demonstrated how these SOS optimization programs can be adapted for rigid-body systems with input limits and hybrid systems with contacts. In future work, we plan to extend the algorithm to more complex systems and deal with disturbances, model uncertainties, and sensor noise.





\section*{ACKNOWLEDGMENT} 
This work was supported by ONR N00014-22-1-2121 and Amazon PO 2D-06310236. The authors would like to thank Jack Umenberger, Kaiqing Zhang and Tobia  Marcucci for valuable discussions and help on the paper. 
\bibliography{root.bib}
\bibliographystyle{IEEEtran}
\section{APPENDIX}
\subsection{Optimal controller with input constraints}
\label{subsection:optimal_controller_proof}
Since $R=\diag(r_1, \cdots, r_{n_u})$ is diagonal, the quadratic program (QP) \eqref{eq:controller_qp} can be decomposed into $n_u$ small scalar QPs and solved separately 
\begin{align}
    &\text{for } m = 1, \cdots, n_u: \nonumber \\
    &u_m^* = \argmin_{u_{\text{min}} \le u_m \le u_{\text{max}}} r_m u_m^2 + \left[\frac{\partial J}{\partial x} f_2(x)\right]_m u_m,
\end{align}
where the subscript $m$ denotes the $m$-th element of a vector. The optimal solution to this single-variable quadratic cost with interval bounds occurs either at the boundary of the interval, or when the gradient of the quadratic cost is zero, whichever takes smaller quadratic cost. Hence in the scalar case, the $m$-th element of the constrained optimal controller is
\begin{align}
    u_m^* = \text{clamp}\left(-\frac{1}{2}r_m^{-1}\left[f_2(x) \frac{\partial{J}}{\partial x}\right]_m, u_{\text{min}}, u_{\text{max}}\right).
\end{align}
\subsection{SOS program}
\label{subsection:sos_program}
For a rigid-body system with $s$ and $c$ (defined in Sec. \ref{subsection:rigid_body_system} Example 1) in the state variable, assume that $\mathcal{X}^{\text{h}}$ can be written as a basic semialgebraic set $\mathcal{X}^{\text{h}} = \{x|g_i(x)\le 0, \forall i = 1, \dots, n\}$ with $g_i(x)$ polynomials in $x$ and the control input set $\mathcal{U} =  \{u|h_j(u)\le 0, \forall j = 1, \dots, m\}$ with $h_j(u)$ polynomials in $u$. The value function under-approximation SOS program is
\begin{align}
&\max_{\underline{J}(x)\succeq 0, \tau(x, u), \bar \tau_i(x, u), \Tilde \tau_j(x, u)} \int_{\mathcal{X}} \underline{J}(x)\; dx  \nonumber\\
\text{ s.t. } & l(x, u) + \frac{\partial \underline{J}}{\partial x} f(x, u) + \tau(x, u) (s^2 + c^2 - 1) \nonumber \\
& + \sum_{i=1}^n \bar \tau_i(x, u) g_i(x) + \sum_{j=1}^m \Tilde \tau_j(x, u) h_j(u)  \text{ is SOS} \nonumber\\
&\bar\tau_i(x, u), \Tilde \tau_j(x, u) \text{ are SOS} \nonumber\\
&\tau(x, u) \text{ is free polynomial}.
\end{align}

\subsection{Proof of Proposition \ref{prop:lower_bound}}
\label{subsection:proposition1_proof}
\begin{mproof}
Starting from $x(0) \in \mathcal{B}_{\pi^*}^{\text{h}}$, integrate the inequality \eqref{eq: J_lower_inequality} along the trajectory $(x,  \pi^*(x))$
\begin{align*}
    0 &\le \int_0^{\infty} [l(x,  \pi^*(x)) + \underline{\dot{J}}^*(x)] \; dt \\
    &= \int_0^{\infty} l(x,  \pi^*(x)) \; dt + \underline{J}^*(x(\infty)) - \underline{J}^*(x(0))\\
    \underline{J}^*(x(0)) &\le \int_0^{\infty} l(x, \pi^*(x)) = J^* (x(0)),
\end{align*}
where the optimal controller $\pi^*$ leads to $x(\infty) = x_d$ and $\underline{J}^*(x(\infty))=0$.
\end{mproof}
\subsection{Proof of Proposition \ref{prop:upper_connected}}
\label{subsection:proposition3_proof}
\begin{mproof}
We begin by noting that $\hat{\mathcal{B}}_{\bar \pi}^{\bar \rho}$ cannot intersect the boundary of $\mathcal{X}$. Indeed if $x_{0} \in \hat{\mathcal{B}}_{\bar \pi}^{\bar \rho}\cap \partial \mathcal{X}$, then we must have that $\bar{J}^*(x_{0}) < \bar{\rho}$ which implies that $\bar{\rho}$ is not the minimum value on the boundary of $\mathcal{X}$.
Now, seeking a contradiction, suppose $\hat{\mathcal{B}}_{\bar \pi}^{\bar \rho}$ is disconnected and thus the disjoint union of a non-empty open set $\mathcal{A}$ and an open set $\mathcal{Z}$ containing $x_d$. Neither of these sets intersect the boundary and therefore $\mathcal{A}, \mathcal{Z} \subset \mathcal{X}$ and so both are bounded. Without loss of generality, we have that the closure $\textbf{cl}(\mathcal{A})$ is a compact set that does not contain $x_{d}$. 

Recall that the stage cost $l(x, u) > 0, ~\forall u \in \mathcal{U}, \forall x \in \mathbb{R}^{n_x} \backslash \{x_d\}$. Moreover, $\mathcal{A}$ is an open sublevel set of $\bar{J}^{*}(x)$ and since $\mathcal{A} \subset \mathcal{X}$, $\dot{\bar{J}}^{*}(x) \leq -l(x,\pi_0(x)) < 0, ~\forall x \in \mathcal{A}$. Therefore, $\mathcal{A}$ is invariant.

By the extreme value theorem, the minimum of $\bar{J}^{*}(x)$ over the compact set $\textbf{cl}(\mathcal{A})$ is attained at some point $x_1$ in $\textbf{cl}(\mathcal{A})$, with $\bar{J}^{*}(x_1)$ strictly less than $\bar \rho$. Since all points $x \in \partial \mathcal{A}$ achieve $\bar{J}^{*}(x) = \bar{\rho}$, we have that $x_1 \in \mathcal{A}$.


 However, $\dot{\bar{J}}^{*}(x_1) < 0$ and so there exists $\hat{x}_{1}$ such that $\bar{J}^{*}(\hat{x}_{1}) < \bar{J}^{*}(x_1)$. Since $\mathcal{A}$ is invariant, $\hat{x}_{1} \in \mathcal{A}$ which contradicts the minimality of $x_1$.

\end{mproof}
\end{document}